\documentclass[runningheads]{llncs}
\usepackage{tabularx}
\usepackage{graphicx}
\usepackage{subfigure}
\usepackage{float}
\usepackage{xcolor}
\usepackage{amsmath}
\usepackage{multirow}
\usepackage{caption}

\begin{document}

\title{A Machine Learning Approach to Analyze the Effects of Alzheimer's Disease on Handwriting through Lognormal Features}

\author {Tiziana D'Alessandro\thanks{Corresponding author (email: \email{tiziana.dalessa@unicas.it}).}\inst{2} \and Cristina Carmona-Duarte\inst{1} \and   Claudio De Stefano\inst{2} \and Moises Diaz\inst{1} \and Miguel Ferrer\inst{1} \and Francesco Fontanella\inst{2}
}
\institute{
IDeTIC: Instituto para el Desarrollo Tecnológico y la Innovación en Comunicaciones. Universidad de Las Palmas de Gran Canaria, Spain
\and
Department of Electrical and Information Engineering (DIEI), University of Cassino and Southern Lazio,
Via G. Di Biasio 43, 03043 Cassino (FR), Italy
}

\maketitle
\begin{abstract}
Alzheimer's disease is one of the most incisive illnesses among the neurodegenerative ones, and it causes a progressive decline in cognitive abilities that, in the worst cases, becomes severe enough to interfere with daily life. Currently, there is no cure, so an early diagnosis is strongly needed to try and slow its progression through medical treatments. Handwriting analysis is considered a potential tool for detecting and understanding certain neurological conditions, including Alzheimer's disease. While handwriting analysis alone cannot provide a definitive diagnosis of Alzheimer's, it may offer some insights and be used for a comprehensive assessment. The Sigma-lognormal model is conceived for movement analysis and can also be applied to handwriting. This model returns a set of lognormal parameters as output, which forms the basis for the computation of novel and significant features. 
This paper presents a machine learning approach applied to handwriting features extracted through the sigma-lognormal model. The aim is to develop a support system to help doctors in the diagnosis and study of Alzheimer, evaluate the effectiveness of the extracted features and finally study the relation among them.
\end{abstract}


\section{Introduction} 
\label{intro}
Neurodegenerative diseases are characterized by the progressive degeneration of neurons in the brain; they are irreversible and can affect cognitive and physical abilities. Due to the life-lengthening, they are becoming increasingly common, but one of them has a higher incidence: Alzheimer's disease (AD).
AD provokes a progressive decline in mental functions, affecting several skills, such as memory, thought, judgment and learning. 
Currently, there is no resolution cure, but medical treatments can help manage the symptoms and slow the progression of AD, improving the quality of life of patients and their family. The effectiveness of those medications is not the same for every person, and it is potentially correlated to the time AD is first detected. This means that the time when the diagnosis is made is crucial; the earliest, the better. These reasons led researchers to continuously investigate new techniques and methods to take care of as early as possible. 
Handwriting is known to be one of the first skills to suffer impairment because of AD, as it is the result of both cognitive and movement skills, and it also requires spatial organization and good coordination. Therefore, the study of handwriting can provide a cheap and completely non-invasive way to evaluate the AD insurgence or progression \cite{IPV18,Study21,Rosenblum}. 
Machine learning (ML) is a subcategory of artificial intelligence that focuses on developing algorithms and models that allow computers to learn from data and make predictions without being explicitly programmed. In recent years, the application of this technology has seen widespread adoption across various fields, including handwriting analysis, motor function rehabilitation, and advancements in the field of medicine \cite{MOL20,Albu19}. The wide spread of technologies aimed to record movements and their kinematics, such as tablets or watches, led researchers to support the diagnosis and the treatment of AD \cite{Tanveer20,V019,IP18,Study23,Study22}. 
In this research, we analysed the handwriting through the Kinematic Theory of rapid movements by applying the Sigma-Lognormal model, for which every complex movement can be decomposed  into a vector summation of simple time-overlapped movements \cite{Plamondon1,PlamondonII,PlamondonIII}. This theory can be applied in many fields for movement modelling, such as speech \cite{Voz}, but also handwriting \cite{xZero1,PlamondonIV,children} and neuromuscular disorders \cite{O'Reilly2012787,Lognomal_Apps,impedovo2019velocity,impedovo2021investigating}. \\
This research aims to develop a classification system for AD diagnosis based on handwriting features extracted by applying the sigma-lognormal model, extending the set computed in \cite{IGS21}. We study the effectiveness of those features through a set of ML experiments detailed in the following sections and discuss the obtained results.\\
The paper is organized as follows: Sec. \ref{sec:model} describes the Sigma-lognormal model, in Sec. \ref{sec:data}, we present the tasks used to collect handwriting data and the features extracted using the Sigma-lognormal model. Sec. \ref{sec:exp} details the experimental workflow, its results and features findings. Sec. \ref{sec:conc} outlines concluding remarks and possible future investigations.

\section{The Sigma-Lognormal Model} 
\label{sec:model}
The Sigma-Lognormal model allows the decomposition of rapid movements into a vector summation of simple time-overlapped movements and focuses on the Kinematic Theory of rapid movements \cite{Plamondon1,PlamondonII,PlamondonIII}. This assumption led to the development of several algorithms, and in this work, we adopted the IDeLog algorithm \cite{Idelog20}.

The Sigma-Lognormal model characterizes the resulting velocity of each individual fast movement primitive by utilizing a lognormal function, which represents the velocity peaks situated between two minimum speeds, effectively modelling the velocity profile:

\begin{equation}
v_j (t;t_{0_j} ,\mu_j,\sigma_j^2 )=
D_j \Lambda (t;t_{0_j} ,\mu_j,\sigma_j^2 )=\frac{D_j}{\sigma_j\sqrt{2\pi}(t-t_{0_j})}exp\{\frac{[-ln(t-t_{0_j})-\mu_j]^2}{2\sigma_j^2}\}
\end{equation}

where $t$ is the time basis, \(D_j\) the amplitude, \(t_{o_j}\) the time of occurrence, \(\mu_j\) the log-time delay and \(\sigma_j\) the response log-time.
The lognormal parameters, \(t_{0_j}\), \(\mu_j\) and \(\sigma_j^2\) are calculated and adjusted by iterative interactions between the original trajectory profile and the reconstructed one to minimize the error between the reconstructed lognormal and the original velocity profile. 

\begin{figure}[t] 
\includegraphics[width=\textwidth]{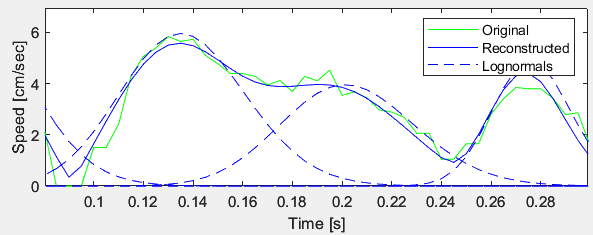}
\caption{Comparison between the original (green) and the reconstructed (blue) velocity profile. Dotted blue lines show the lognormal functions that generated the reconstructed profile.}
\label{fig:log}
\end{figure}

Fig. \ref{fig:log}, illustrates a case of a rapid movement, and as it is the succession of $M$ simple movements, its velocity profile is obtained from the time superposition of $M$ lognormals:

\begin{equation}
v_n(t)=\sum_{j=1}^{M}v_j (t;{t_{0_j}} ,\mu_j,\sigma_j^2)=\sum_{j=1}^{M}D_j
\begin{bmatrix}
\cos(\Phi_j (t))\\
\sin(\Phi_j(t))
\end{bmatrix}
\Lambda (t;t_{0_j} ,\mu_j,\sigma_j^2 )
\end{equation}
where  \(\Phi_j(t)\) is the angular position given by:
\begin{equation}
\Phi_j (t)=\Theta_{s_j}+\frac{(\Theta_{e_j}-\Theta_{s_j})}{2}[1+erf(\frac{\ln(t-t_{0_j})-\mu_j}{\sigma_j\sqrt{2}})]
\end{equation}
being \(\Theta_{s_j}\) and  \(\Theta_e{_j}\) are the starting and the end angular direction of  the \(j^th\) simple movement. 

\section{Description}
\label{sec:description}
This Section details the handwriting tasks considered, how they were acquired and the features extracted through the sigma-lognormal model.

\subsection{Data collection}
\label{sec:data}
The data collection comes from the execution of a protocol \cite{CDF18} composed of different kinds of handwriting tasks. They were performed with the Wacom Bamboo Folio tablet. Such a tool allows recording the handwriting in terms of $x-y$ spatial coordinates and pressure $p$, for each point at a sampling rate of 200Hz. A total of 174 people were involved in the acquisition phase, 89 patients (PT) suffering from AD and a healthy control group (HC) of 85 people. Participants were selected  with the support of the geriatrics department and Alzheimer's unit of the "Federico II" hospital in Naples. The recruiting criteria were based on standard clinical tests, such as the Mini-Mental State Examination (MMSE), the Frontal Assessment Battery (FAB) and the Montreal Cognitive Assessment (MoCA). 

\subsection{Tasks}
This research aims to check which task performs best with our system and understand if the sigma-lognormal model and the extracted features are adequate for our problem.  This reasoning led us to consider the data collected from each task execution to test our experimental setting. In \cite{CDF18} is detailed the protocol used and the 25 tasks that made it. The choice of tasks was based on literature to analyze different aspects of handwriting and the deterioration of the skills required to perform them. Every task requires different abilities that AD may compromise, like cognitive, kinesthetic and perceptive-motor functions \cite{Tseng93}, such as language comprehension, muscle control, spatial organization and coordination.
Four different groups of tasks can be distinguished in the protocol, taking into account their objectives:
\begin{itemize}
    \item Graphic tasks to test the patient’s ability in writing elementary traits, joining some points and drawing figures (simple or complex and scaled in various dimensions).
    \item Copy and Reverse Copy tasks, to test the patient’s abilities in repeating complex graphic gestures with a semantic meaning, such as letters, words and numbers (of different lengths and with different spatial organizations).
    \item Memory tasks, to test the variation of the graphic section, keep a word, a letter, a graphic gesture or a motor planning in memory.
    \item Dictation, to investigate how the writing in the task varies (with phrases or numbers) in which the use of the working memory is necessary.
\end{itemize}
There is a rationale behind the choice of every subgroup of tasks that comes from the study of the symptoms of Alzheimer's \cite{CDF18}. It's known that this disease's effects can change from person to person. Some people show more impairment on the mental side, others on the muscular side, and many people can also find compensation for them, mostly for the physical ones. With its 25 tasks, this protocol means to study if the handwriting is altered by Alzheimer's, taking into account its symptoms.

\subsection{Feature Engineering}
\label{sec:features}
The feature engineering process was done taking into account the outcomes from \cite{PlamondonIV,children}, but also several studies about the normative range of variations in the lognormal parameters, which give a notion of how an ideal movement could be \cite{lognormal}, based on lognormal movement decomposition. In this way, it was possible to extract a wide set of features that could be exhaustive to characterize one person's handwriting and, in particular, to enhance an eventual difference between the execution of a person affected by Alzheimer's and healthy control. The first step of this process is applying the Sigma-Lognormal model, defined in Sec. \ref{sec:model}, to the data acquired as mentioned in Sec. \ref{sec:data}. As a result, every task was decomposed into a vector summation of simple time-overlapped movements, for each one is associated with a lognormal function and a set of Sigma-Lognormal parameters was obtained \(P_j=[D_j, t_{0_j}, \mu_j, \sigma_j, \theta_{s_j}, \theta_{e_j}]\), where $j$ refers to the $jth$ lognormal. From each task execution, we only processed points from the first time the pen touched the paper to the last time (first and last pen-down), as the tool also record movements when the person is approaching for the first time to the the paper or is leaving it for the last time. This step was necessary to clean the data from those movements that don't belong to the execution of the task but precede or follow the real handwriting gesture we meant to analyze. 
Once obtained the sigma-lognormal parameters three groups of features were computed, related to different aspects and measures of the handwriting execution:

\begin{enumerate}
    \item Time: features that represent temporal aspects of the execution, such as total time to execute a task, contact time, that is, the portion of total time in which the movements were performed without losing contact with the tablet, it means that the pen was at a maximum distance of 3 cm from the tablet surface. The remaining portion of the total time is the losing time. Some of these features are also related to the number of lognormals counted in the reconstructed velocity profile. This information is usually proportionally related to the task and the time of execution;
    \item Signal-to-noise ratio (SNR): this measure, and features related to it, give the information of the goodness of the reconstructed trace (SNRt) and velocity profile (SNRv) from the sigma-lognormal model;
    \item Geometric shapes of the reconstructed speed profile: they are useful to understand the movements' velocity, stability and fluency. Some features are computed starting from the lognormal parameters $D$ and $\sigma$; others are from geometrical shapes (area, height and width) of lognormals in the reconstructed velocity profile. In detail, with the term area, we refer to the overlapping area between two consecutive lognormal, while height is the maximum and width is the base of a lognormal function \cite{children}.
\end{enumerate}

Tables \ref{tab:features1} and \ref{tab:features2} show the computed features, whose name is written in the form of $fxx$ next to their explanation. 
This step aims to understand if AD can be estimated using the features extracted from handwriting movements through the sigma-lognormal model. Besides the aforementioned features group, we also used personal features in the experiments: age, gender, education and type of profession. Alzheimer's effects lead to the choice of every group of features. Temporal features are interesting because a person who is impaired should take more time to execute a task and, according to the illness progression level, also more losing time. The latter is the amount of time the person lifted the pen too far from the tablet to be detected, maybe because of fatigue or distraction. Among the time features, there is also the number of lognormal generated from the velocity profile and the number of segments, where each segment refers to an entire trace acquired without losing contact. We expect all the temporal features to be higher for people affected by AD.
The signal-to-noise ratio measures the reconstruction quality and, when divided by the number of lognormals is useful to describe how fluent a movement is \cite{PlamondonIV}.
Geometrical features are computed starting from three sequences: overlapping areas, heights and widths of the lognormal functions. They give us information about the fluency of the handwriting: the greater the overlapping area, the more fluent the handwriting, without strong deceleration or pauses. The height is proportional to the speed, while a larger width denotes a slower movement. Among the geometrical features, some related to the lognormal parameters D and sigma give information about the lognormal distance covered in the kinematic space and the lognormal response time.
Understanding how those measures change during a handwriting task or relating them to the temporal feature can provide valuable information.

\begin{table}[htbp]
  \centering
  \caption{Time and SNR related features}
    \begin{tabular}{l|l|c|l}
    \multicolumn{4}{c}{\textbf{Features}} \\
    \hline
    \multicolumn{2}{c}{\textbf{TIME}} & \multicolumn{2}{c}{\textbf{SNR}} \\
    \hline
    f1    &  number of lognormals & f15   & mean(SNRt) \\
    f2    &  number of segments & f16   & std(SNRt) \\
    f3    & task total time & f17   & mean(SNRv) \\
    f4    &  contact time & f18   & std(SNRv) \\
    f5    & losing time & f19   & sum(SNRt)/f1 \\
    f6    & standard deviation of seg. time & f20   & f15/f1 \\
    f7    & f3/f2 & f21   & f16/f1 \\
    f8    & f3/f1 & f22   & sum(SNRv)/f1 \\
    f9    & f4/f2 & f23   & f17/f1 \\
    f10   & f4/f1 & f24   & f18/f1 \\
    f11   & f5/f2 &       &  \\
    f12   & f5/f1 &       &  \\
    f13   & mean(number of log.s per seg.) &       &  \\
    f14   & std(number of log.s per seg.) &       &  \\
    \hline
    \end{tabular}%
  \label{tab:features1}%
\end{table}%

\begin{table}[htbp]
  \centering
  \caption{Geometrical features}
    \begin{tabular}{c|l|c|l|c|l}
    \multicolumn{6}{c}{\textbf{Features}} \\
    \hline
    \multicolumn{6}{c}{\textbf{GEOMETRICAL}} \\
    \hline
    f25   & std(areas) & f39   & mean(areas)/f25 & f53   & dif(widths)/1 \\
    f26   &  std(heights) & f40   & mean(heights)*exp(f26) & f54   & dev(widths)/1 \\
    f27   & std(widths) & f41   & mean(heights)*ln(f26) & f55   &  'seg\_difA\_div\_nlog' \\
    f28   & sum(areas)/f3 & f42   & mean(heights)/f26 & f56   & 'std\_seg\_difA\_div\_nlog' \\
    f29   & sum(areas)/f4 & f43   & mean(widths)*exp(f27) & f57   & dif(sigma)/f1 \\
    f30   & sum(areas)/f1 & f44   & mean(widths)*ln(f27) & f58   & std(sigma)/f1 \\
    f31   & sum(heights)/f3 & f45   & mean(widths)/f27 & f59   & dif(sigma)/f4 \\
    f32   & sum(heights)/f4 & f46   & f25/f4 & f60   & std(sigma)/f4 \\
    f33   & sum(heights)/f1 & f47   & f26/f4 & f61   & dif(D)/f1 \\
    f34   & sum(widths)/f3 & f48   & f27/f4 & f62   & std(D)/f1 \\
    f35   & sum(widths)/f4 & f49   & dif(areas)/f1 & f63   & dif(D)/f4 \\
    f36   & sum(widths)/f1 & f50   & std(areas)/f1 & f64   & std(D)/f4 \\
    f37   & mean(areas)*exp(f25) & f51   & dif(heights)/f1 &       &  \\
    f38   & mean(areas)*ln(f25) & f52   & std(heights)/f1 &       &  \\
    \hline
    \end{tabular}%
  \label{tab:features2}%
\end{table}%

\section{Experimental phase}
\label{sec:exp}
This section describes the experimental phase of the research and the results achieved. The experiments were carried out relying on classical machine learning techniques and algorithms. The following subsection illustrates the adopted workflow. 

\subsection{Workflow}
\label{sec:workflow}
\begin{figure}[t] 
\includegraphics[width=\textwidth]{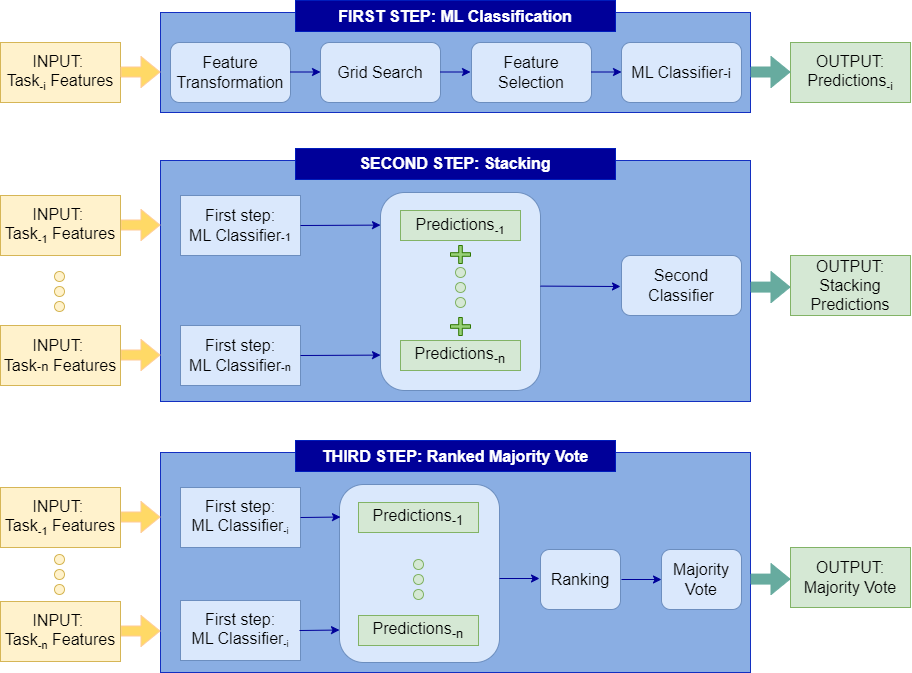}
\caption{Workflow Representation.}
\label{fig:workflow}
\end{figure}

The previous feature extraction step, discussed in Sec. \ref{sec:features}, generated a set of several features for every task, so every task has its dataset. 
The workflow adopted can be better discussed considering a three-step approach, discussed in the following and shown in Fig. \ref{fig:workflow}.

\subsubsection{First Step: ML Classification}
\label{sec:ml_alg}
For the first step, we chose seven well-known classification algorithms to perform a classification for every task dataset: XGBoost (XGB), Random Forest (RF), Decision Tree (DT), Support Vector Machine (SVM), Multilayer Perceptron (MLP), K-Nearest Neighbors (KNN), Logistic Regression (LR). Before proceeding with the training, we applied three ML techniques to increase the discriminative power of our system:
\begin{itemize}
    \item Feature scaling;
    \item Grid search;
    \item Feature Selection: RFECV/SelectKBest.
\end{itemize}

The grid search procedure was carried out with 50\% of data randomly selected from the whole dataset. Once the best set of hyperparameters and features for every algorithm was obtained, the training started one task at a time. In detail, we randomly divided the dataset into train and test sets, assigning to each set 50\% of the total samples, keeping the balance between the two classes of the problem: Healthy Controls (HC) and Patients (PT).
Thirty random runs were performed to obtain more reliable and robust performance estimates for the classifier, and the final results were averaged over the thirty runs.

\subsubsection{Second Step: Stacking}
\label{sec:stacking}
The second step of our classification approach is a stacking technique \cite{Stacking}. It is an ML approach, a stacked generalization or ensemble stacking. It combines the predictions of multiple models or base learners to create a more powerful and accurate final prediction. It involves training multiple diverse models on the same dataset and then using a "meta-learner" to learn from the predictions of these base models. We considered this technique to increase the predictive power and robustness of our system. By combining multiple models, in fact, you can leverage their complementary strengths and reduce their individual weaknesses, potentially capturing complex relationships and patterns in the data. Given these assumptions, we used the output prediction provided by the classifiers of the first step. In particular, we merged the responses obtained for all the tasks so that the new feature vector for each sample (person) comprises the predictions that a classifier has attributed to that sample for each task. As the process is iterated over all the runs, the final score is the average of the stacking results obtained over the 30 runs. 
After a testing phase, we selected XGB as the estimator of the stacking technique, as it is the classifier which allowed us to reach the highest performance.

\subsubsection{Third Step: Ranking and Majority vote}
\label{sec:mv}
The third step of the experimental approach was inputting the outcome from the first step detailed above (Sec. \ref{sec:ml_alg}). First, for each classifier, we performed a ranking technique, an algorithm that orders a set of items, in our case, the tasks, based on their relevance, the accuracy metric. We chose accuracy because it measures the classifier's effectiveness, so we achieved a list of tasks sorted in ascendent order concerning this metric.
This process is followed by a combination rule: the majority vote. This rule is applied for problems with multiple classifiers or models to make predictions, and the majority opinion determines the final prediction. Each classifier's prediction is considered a vote, and the class with the most votes is chosen as the final prediction. We applied it by combining every classifier prediction for a different set of tasks, run by run, and finally, we averaged the accuracy over the thirty runs. We considered the list of tasks given as output by the ranking process to select significant subsets of tasks to apply the majority vote.

\subsection{Results}
\label{sec:res}
The results obtained with the experimental setting, reported in Sec. \ref{sec:workflow}, are shown and discussed in the following.

\subsubsection{First Step Results}
\label{sec:ml_res}
The results obtained from the first experimental step are shown in Table \ref{tab:ml_accuracy}, as the average accuracy (percentage value) for each task and classifier over 30 runs. Note that for every task, the performance of the best classifier is in bold. The average accuracies range from the minimum value of 58.31\%, achieved from KNN on the 1st task, to the maximum of 78.41\% obtained from RF on the 23rd task. In particular, RF outperformed the other classifiers on seven tasks out of 25, while KNN never reached the highest result. Independently from the type of classifier, the table shows that the 1st task obtained the worst performances, while the 23rd was the best. The 1st task required a person to perform their signature; instead, the 23rd required writing a telephone number under dictation.
As the 23rd task reached the best accuracy, we decided to investigate more aspects of this experiment by computing more evaluation metrics \cite{metrics_book}: precision (PRE), sensitivity (SEN), specificity (SPE), False Negative Rate (FNR), F1 score (F1S) and Area Under the Curve (AUC) \cite{AUC}. Table \ref{tab:task23} contains the classifier used on the 23rd task in the first column, while the other columns report the metric's value in the header, averaging over 30 runs. All the metrics are expressed as percentage values except for the AUC, which varies from 0 to 1. We enhanced in bold the best metric value in every column. This table shows that the best classifier for this task was RF, according to the accuracy and the other metrics, except for the precision and the specificity. These last two indicate that RF wasn't the best classifier to classify healthy controls correctly, but in the medical field, the most important thing is to identify those affected by the illness, as a false prediction has more consequences in this case. The FNR is at 22.07\%, and it's the lowest value reached among all the classifiers, meaning that RF was the best to recognize patients.
Table \ref{tab:comparison} shows a comparison between the results obtained on every task using a set of dynamic features, with those obtained with the lognormal features extracted for this study.

The dynamic features are handwriting characteristics, including Start time, Duration, Vertical dimension, Horizontal dimension, Vertical speed peak, Peak of vertical acceleration, Relative initial inclination, Jerk, Pen pressure etc.
A feature vector is obtained for each task performed by each subject. It is worth noticing that those features have been considered in this study for comparison purposes only. An in-depth analysis can be found in \cite{Cilia21b}.

In detail, the table contains accuracy percentage values computed by averaging this metric over 30 runs. Only the results of the best classifier are shown, and the best performance for each task is in bold. Most of the time, the classification with our proposed lognormal features allowed us to reach better results.

\begin{figure}[htbp]
\caption{Average Accuracy achieved on 30 runs for every ML algorithm on lognormal features}
  \begin{minipage}{0.40\linewidth} 
    \centering
    \begin{tabular}{c|ccccccc}
    \multicolumn{8}{c}{\textbf{Accuracy}} \\
    \hline
    \textbf{T} & \textbf{XGB} & \textbf{RF} & \textbf{DT} & \textbf{SVM} & \textbf{MLP} & \textbf{KNN} & \textbf{LR} \\
    \hline
    1     & \textbf{67.6} & 66.3 & 60.3 & 62.7 & 60.1 & 58.3 & 65.0 \\
    2     & 65.3  & 66.3 & 60.0 & \textbf{68.6} & 65.5 & 61.7 & 65.1 \\
    3     & 66.9 & 68.1 & 64.8 & \textbf{68.5} & 63.8 & 62.0 & 67.6 \\
    4     & 65.0 & 66.3 & 58.4 & 66.4 & 66.6  & 62.6  & \textbf{67.7} \\
    5     & 67.2 & 69.0 & 62.0 & 66.8 & 65.0 & 61.5 & \textbf{69.9} \\
    6     & 70.6  & 75.0 & 67.4 & \textbf{75.6} & 61.8 & 64.1 & 75.0 \\
    7     & 70.5 & 68.61 & 68.6  & \textbf{73.7} & 69.8 & 66.3 & 71.4 \\
    8     & 68.3 & 68.6 & 65.2 & \textbf{69.1} & 68.2 & 64.5 & 65.6 \\
    9     & 76.5 & \textbf{77.3} & 70.0 & 74.6 & 67.1 & 74.4 & 76.0 \\
    10    & \textbf{71.2} & 69.5 & 62.8 & 68.9 & 63.5 & 65.0 & 70.2 \\
    11    & 69.0 & 69.3 & 63.3 & 65.5 & 68.0 & 64.7 & \textbf{70.1} \\
    12    & 68.3 & 68.6 & 62.4  & 66.1 & 64.9 & 60.5 & \textbf{70.6} \\
    13    & \textbf{67.3} & 62.6 & 58.7 & 63.7 & 67.1 & 62.8 & 66.7 \\
    \hline
    \end{tabular}%
  \end{minipage}
  \hspace{1.8cm}   
  \begin{minipage}{0.40\linewidth} 
  \centering
    \begin{tabular}{c|ccccccc}
    \multicolumn{8}{c}{\textbf{Accuracy}} \\
    \hline
    \textbf{T} & \textbf{XGB} & \textbf{RF} & \textbf{DT} & \textbf{SVM} & \textbf{MLP} & \textbf{KNN} & \textbf{LR} \\
    \hline
    14    & \textbf{68.3} & 67.4 & 61.6 & 66.2 & 64.9 & 65.0 & 66.7 \\
    15    & 71.0 & 72.5 & 67.6 & 73.2 & 73.0 & 69.3 & \textbf{73.3} \\
    16    & 65.8 & 64.2 & 59.2 & \textbf{67.4} & 63.0 & 61.4 & 67.4 \\
    17    & 74.6 & \textbf{75.7} & 71.3 & 71.8 & 70.9 & 65.6 & 75.0 \\
    18    & 64.5 & \textbf{68.4} & 62.2 & 67.2 & 64.9 & 62.9 & 67.7 \\
    19    & 65.0 & \textbf{66.4} & 61.9 & 66.2 & 59.4 & 66.0 & 65.1 \\
    20    & 66.2 & 66.8 & 64.7 & 67.6 & 66.1 & 66.5 & \textbf{69.9} \\
    21    & 66.3 & \textbf{67.2} & 58.7 & 63.8 & 59.0 & 61.8 & 67.0 \\
    22    & \textbf{72.9} & 72.3 & 68.3 & 71.6 & 68.6 & 66.7 & 68.8 \\
    23    & 77.4 & \textbf{78.4} & 70.7 & 78.3 & 66.7 & 75.0 & 78.0 \\
    24    & \textbf{76.4} & 73.9 & 65.9 & 65.3 & 68.0 & 62.3 & 67.5 \\
    25    & 72.8 & \textbf{74.6} & 63.2 & 73.1 & 68.8 & 68.1 & 71.3 \\
        &  &  &  &  &  &  &  \\
    \hline
    \end{tabular}%
  \end{minipage}
  \label{tab:ml_accuracy}
\end{figure}

\begin{table}[htbp]
  \centering
  \caption{Average results achieved on 30 runs for every ML algorithm on lognormal features, extracted from the execution of task 23, ie the one that reached the best performance according to Table \ref{tab:ml_accuracy}}
    \begin{tabular}{l|rrrrrrr}
    \hline
    \textbf{CLS} & \multicolumn{1}{l}{\textbf{ACC}} & \multicolumn{1}{l}{\textbf{PRE}} & \multicolumn{1}{l}{\textbf{SEN}} & \multicolumn{1}{l}{\textbf{SPE}} & \multicolumn{1}{l}{\textbf{FNR}} & \multicolumn{1}{l}{\textbf{F1S}} & \multicolumn{1}{l}{\textbf{AUC}} \\
    \hline
    XGB   & 77.43  & 77.67 & 77.03 & 77.88 & 22.97 & 77.18 & 0.83 \\
    RF    & \textbf{78.41}  & 78.72 & \textbf{77.93} & 78.89 & \textbf{22.07} & \textbf{78.14} & \textbf{0.85} \\
    DT    & 70.78  & 78.77 & 58.16 & 83.25 & 41.84 & 65.21 & 0.70 \\
    SVM   & 78.39  & 81.53 & 73.51 & 83.23 & 26.49 & 77.06 & 0.84 \\
    MLP   & 66.72  & \textbf{90.23} & 36.38 & \textbf{96.73} & 63.62 & 49.67 & 0.81 \\
    KNN   & 75.06  & 77.04 & 71.57 & 78.54 & 28.43 & 74.01 & 0.80 \\
    LR    & 78.02  & 79.88 & 75.37 & 80.69 & 24.63 & 77.24 & 0.84 \\
    \hline
    \end{tabular}%
  \label{tab:task23}%
\end{table}%

\begin{figure}[htbp]
\caption{Comparison between average Accuracy achieved on 30 runs for every task with the best-performing ML algorithm for Dynamic and Lognormal features}
  \begin{minipage}{0.40\linewidth} 
    \centering
    \begin{tabular}{c|l|c|l|c}
    \hline
          & \multicolumn{2}{c|}{\textbf{DYN. FEAT.}} & \multicolumn{2}{c}{\textbf{LOG. FEAT.}} \\
    \hline
    \textbf{T} & \multicolumn{1}{c|}{\textbf{CLS}} & \textbf{ACC} & \multicolumn{1}{c|}{\textbf{CLS}} & \textbf{ACC} \\
    \hline
    1     & XGB   & 64.5 & XGB   & \textbf{67.6} \\
    2     & XGB    & 63.0 & SVM   & \textbf{68.6} \\
    3     & XGB    & 57.3 & SVM   & \textbf{68.5} \\
    4     & XGB   & 59.2 & LR    & \textbf{67.7} \\
    5     & DT   & 69.0 & LR    & \textbf{69.9} \\
    6     & XGB    & 57.4 & SVM   & \textbf{75.6} \\
    7     & DT   & 58.4 & SVM   & \textbf{73.7} \\
    8     & DT   & 61.7 & SVM   & \textbf{69.1} \\
    9     & XGB   & 64.5 & RF    & \textbf{77.3} \\
    10    & XGB   & 61.6 & XGB   & \textbf{71.2} \\
    11    & XGB   & 66.4 & LR    & \textbf{70.1} \\
    12    & XGB   & 67.2 & LR    & \textbf{70.6} \\
    13    & XGB   & \textbf{68.3} & XGB   & 67.3 \\
    \hline
    \end{tabular}%
  \end{minipage}
  \hspace{1.50cm}   
  \begin{minipage}{0.40\linewidth} 
  \centering
    \begin{tabular}{c|r|c|r|c}
    \hline
          & \multicolumn{2}{c|}{\textbf{DYN. FEAT.}} & \multicolumn{2}{c}{\textbf{LOG. FEAT.}} \\
    \hline
    \textbf{T} & \multicolumn{1}{c|}{\textbf{CLS}} & \textbf{ACC} & \multicolumn{1}{c|}{\textbf{CLS}} & \textbf{ACC} \\
    \hline
    14    & \multicolumn{1}{l|}{XGB} & 64.1 & \multicolumn{1}{l|}{XGB} & \textbf{68.3} \\
    15    & \multicolumn{1}{l|}{XGB} & 64.3 & \multicolumn{1}{l|}{LR} & \textbf{73.3} \\
    16    & \multicolumn{1}{l|}{XGB} & 67.0 & \multicolumn{1}{l|}{SVM} & \textbf{67.4} \\
    17    & \multicolumn{1}{l|}{XGB} & 69.3 & \multicolumn{1}{l|}{RF} & \textbf{75.7} \\
    18    & \multicolumn{1}{l|}{XGB} & \textbf{68.9} & \multicolumn{1}{l|}{RF} & 68.4 \\
    19    & \multicolumn{1}{l|}{XGB} & \textbf{68.3} & \multicolumn{1}{l|}{RF} & 66.4 \\
    20    & \multicolumn{1}{l|}{XGB} & 66.1 & \multicolumn{1}{l|}{LR} & \textbf{69.9} \\
    21    & \multicolumn{1}{l|}{DT} & 55.1    & \multicolumn{1}{l|}{RF} & \textbf{67.2} \\
    22    & \multicolumn{1}{l|}{RF} & 70.0 & \multicolumn{1}{l|}{XGB} & \textbf{72.9} \\
    23    & \multicolumn{1}{l|}{XGB} & 72.3 & \multicolumn{1}{l|}{RF} & \textbf{78.4} \\
    24    & \multicolumn{1}{l|}{XGB} & 56.0 & \multicolumn{1}{l|}{XGB} & \textbf{76.4} \\
    25    & \multicolumn{1}{l|}{DT} & 59.8 & \multicolumn{1}{l|}{RF} & \textbf{74.6} \\
          &       &       &       &  \\
    \hline
    \end{tabular}%
  \end{minipage}
  \label{tab:comparison}
\end{figure}

\subsubsection{Second Step Results}
\label{sec:stack_res}
As mentioned above, stacking is a very popular technique in ML when dealing with multiple classifiers because it can potentially improve the overall predictive performance compared to using individual models separately. The meta-model can learn to leverage the strengths of different base models and compensate for their weaknesses. 
Table \ref{tab:stacking} displays the evaluation for every classifier according to different metrics: accuracy (ACC), precision (PRE), sensitivity (SEN), specificity (SPE), False Negative Rate (FNR), F1 score (F1S) and Area Under the Curve (AUC). Those parameters allowed us to analyze a classifier's performance better. Regarding the accuracy, for each classifier, the stacking reached a value higher than the average accuracy over all tasks from the first step of classification. Considering all the parameters, the best result is given by applying the stacking on the outcome from XGB as the first classifier, and the final stacking accuracy is 76.29\%.

\begin{table}[htbp]
  \centering
  \caption{Stacking results averaged over thirty runs with XGB classifiers, with the output of first-step classifiers}
    \begin{tabular}{l|ccccccc}
    \hline
    \textbf{1st CLS} & \multicolumn{1}{l}{\textbf{ACC}} & \multicolumn{1}{l}{\textbf{PRE}} & \multicolumn{1}{l}{\textbf{SEN}} & \multicolumn{1}{l}{\textbf{SPE}} & \multicolumn{1}{l}{\textbf{FNR}} & \multicolumn{1}{l}{\textbf{F1S}} & \multicolumn{1}{l}{\textbf{AUC}} \\
    \hline
    XGB   & \textbf{76.29} & 77.99 & \textbf{76.09} & 76.52 & \textbf{23.91} & \textbf{76.32} & \textbf{0.84} \\
    RF    & 75.15 & 76.78 & 74.55 & 75.75 & 25.45 & 75.31 & 0.83 \\
    DT    & 70.98 & 73.89 & 68.85 & 73.39 & 31.15 & 70.2  & 0.78 \\
    SVM   & 75.38 & \textbf{78.45} & 72.24 & \textbf{78.7}  & 27.76 & 74.91 & 0.83 \\
    MLP   & 69.41 & 70.78 & 69.86 & 69.00    & 30.14 & 69.63 & 0.76 \\
    KNN   & 71.75 & 74.79 & 69.1  & 74.58 & 30.9  & 71.16 & 0.77 \\
    LR    & 75.68 & 76.78 & 76.02 & 75.3  & 23.98 & 75.81 & 0.83 \\
    \hline
    \end{tabular}%
  \label{tab:stacking}%
\end{table}%

\subsubsection{Third Step Results}
\label{sec:mv_res}
The third step of our experimental phase, as described in Sec. \ref{sec:mv}, regards the application of two popular techniques in ML: ranking and Majority vote. Table \ref{tab:ranking} shows the ranked lists of tasks based on their accuracy for each classifier. Looking at the table is easy to notice the multiple occurrences of some tasks in the first positions, independently from the algorithm. In detail, task number 23 occurs at the first position five times out of seven algorithms; task 9 is between the first three positions for six algorithms, and task 17 is in the first four positions for five algorithms. Other tasks that also seem to show relevance are 6, 7, 15, 22 and 24. The 23rd required the writing of a phone number under dictation, the 9th the writing of the bigram 'le' four times continuously, and for the 17th the person had to write 6 words in defined boxes; every word had a different level of complexity. Regarding the other relevant tasks, the 6th involves the writing of 'l, m, p'; the 7th 'n, l, o, g' in apposite spaces; the 15th is a reverse copy of 'bottiglia'; the 22nd the direct copy of a phone number; while the 24 is the clock drawing test.  
Table \ref{tab:mv} shows the performance of the majority vote for different sets of tasks taking the first n tasks from the ranking lists. The first column, denoted as "T\_set", represents the number of tasks considered for each set of tasks, ranging from a minimum of three to a maximum of 25, which means all the tasks. After the fifth set, 11 tasks, the majority vote accuracy decreases. The best majority vote accuracy is 82.5\% achieved by combining the predictions obtained by XGB from the first three tasks of the ranked list.

\begin{table}[htbp]
  \centering
  \caption{Tasks ranking for each ML Classifier}
    \begin{tabular}{ccccccc}
    \hline
    \textbf{XGB} & \textbf{RF} & \textbf{DT} & \textbf{SVM} & \textbf{MLP} & \textbf{KNN} & \textbf{LR} \\
    \hline
    T23   & T23   & T17   & T23   & T15   & T23   & T23 \\
    T09   & T09   & T23   & T06   & T17   & T09   & T09 \\
    T24   & T17   & T09   & T09   & T07   & T15   & T06 \\
    T17   & T06   & T07   & T07   & T25   & T25   & T17 \\
    T22   & T25   & T22   & T15   & T22   & T22   & T15 \\
    T25   & T24   & T15   & T25   & T08   & T20   & T07 \\
    T10   & T15   & T06   & T17   & T24   & T07   & T25 \\
    T15   & T22   & T24   & T22   & T11   & T19   & T12 \\
    T06   & T10   & T08   & T08   & T13   & T17   & T10 \\
    T07   & T11   & T03   & T10   & T09   & T14   & T11 \\
    T11   & T05   & T20   & T02   & T23   & T10   & T20 \\
    T12   & T12   & T11   & T03   & T04   & T11   & T05 \\
    T08   & T08   & T25   & T20   & T20   & T08   & T22 \\
    T14   & T07   & T10   & T16   & T02   & T06   & T18 \\
    T01   & T18   & T12   & T18   & T05   & T18   & T04 \\
    T13   & T03   & T18   & T05   & T14   & T13   & T03 \\
    T05   & T14   & T05   & T04   & T18   & T04   & T24 \\
    T03   & T21   & T19   & T19   & T12   & T24   & T16 \\
    T21   & T20   & T14   & T14   & T03   & T03   & T21 \\
    T20   & T19   & T01   & T12   & T10   & T21   & T13 \\
    T16   & T02   & T02   & T11   & T16   & T02   & T14 \\
    T02   & T01   & T16   & T24   & T06   & T05   & T08 \\
    T19   & T04   & T13   & T21   & T01   & T16   & T19 \\
    T04   & T16   & T21   & T13   & T19   & T12   & T02 \\
    T18   & T13   & T04   & T01   & T21   & T01   & T01 \\
    \hline
    \end{tabular}%
  \label{tab:ranking}%
\end{table}%

\begin{table}[htbp]
  \centering
  \caption{Majority vote to a different set of ranked tasks.}
  
    \begin{tabular}{c|ccccccc}
    \hline
    \textbf{T\_set} & \textbf{XGB} & \textbf{RF} & \textbf{DT} & \textbf{SVM} & \textbf{MLP} & \textbf{KNN} & \textbf{LR} \\
    \hline
    3     & \textbf{82.5} & 79.23 & 76.16 & 81.22 & \textbf{76.32} & \textbf{77.40} & 79.35 \\
    5     & 79.81 & 80.35 & 76.10  & \textbf{81.88} & 75.37 & 75.21 & \textbf{79.95} \\
    7     & 79.46 & \textbf{81.17} & 77.30  & 80.26 & 74.53 & 75.06 & 79.72 \\
    9     & 79.67 & 80.10  & \textbf{77.75} & 79.99 & 74.37 & 74.99 & 78.10\\
    11    & 78.82 & 79.70  & 77.58 & 80.32 & 75.54 & 74.84 & 78.02 \\
    13    & 77.94 & 78.35 & 77.14 & 78.64 & 75.06 & 74.33 & 77.67 \\
    15    & 77.86 & 78.08 & 76.80  & 77.76 & 73.60  & 72.66 & 76.42 \\
    17    & 77.60  & 77.24 & 76.92 & 77.20  & 72.43 & 72.71 & 76.69 \\
    19    & 77.19 & 77.75 & 76.37 & 76.83 & 72.23 & 72.37 & 76.36 \\
    21    & 76.74 & 77.34 & 75.42 & 77.24 & 72.51 & 72.58 & 75.56 \\
    23    & 76.54 & 77.07 & 75.53 & 77.49 & 71.73 & 72.04 & 74.58 \\
    \hline
    \textbf{ALL}   & 76.39 & 76.64 & 75.19 & 76.99 & 71.93 & 71.76 & 74.04 \\
    \hline
    \end{tabular}%
  \label{tab:mv}%
\end{table}%

\subsection{Feature Findings}
\label{sec:features_res}
This Section is conceived for the discussion of findings related to the feature selection output and the relation between the computed features and a person's age and educational level.

\subsubsection{Feature Selection Discussion}
\label{sec:feature_sel}
We applied a feature selection procedure after the grid search in the first step of the experiment phase. This allowed the classifier to concentrate on the most important features by deleting redundant or uninteresting ones. The chosen algorithm was Recursive Feature Elimination (RFECV) for most classifiers, except for MLP and KNN, for which we used SelectKBest. Different sets of features were selected for every classifier and task, but looking at them, it was possible to notice some common patterns and make some considerations.
Some personal features were always selected for almost every task: age, profession and education.
Regarding temporal features the most selected were f1, f4, f3, f7 and f9; while, among the geometrical ones the most selected were f62, f47, f26, f28, f61, f60, f49, f64; finally for SNR features f19 f20 f22 f23 were selected the most. All these features have been selected on at least ten tasks out of 25 from the classifier that achieved the best result, XGB. 
The features selected from the temporal set denote that our assumptions had foundations: impaired people take more time to execute a handwriting task, generating more lognormal functions in the velocity profile and segments in the trace acquisition. Among features related to SNR and geometric shapes, the most important are those that describe the variation of a measure sequence. In particular, relating SNR or geometrical features to the temporal ones can help to make more robust features enlarging the difference between the two classes we mean to distinguish.\\ 
To better understand the relevance of each feature for the examined problem, a parametric statistical test, the $t-test$, was used to evaluate whether the difference between the means of the two groups was statistically significant.
This test returns a probability measure called the $p-value$ for every characteristic that indicates the strength of evidence against the null hypothesis.
A significance level of 0.05 is the predetermined threshold below which the null hypothesis is rejected, so if the $p-value$ is smaller than this threshold, there is evidence that the feature is significant in distinguishing between the two classes. All the aforementioned features reported a $p-value$ smaller than the threshold in our case.

\subsubsection{Relations between features, educational level and age}
\label{sec:feat_rel}
\begin{figure*}[t]
\centering
\subfigure[All, first age range]{\centering\includegraphics[scale=0.37]{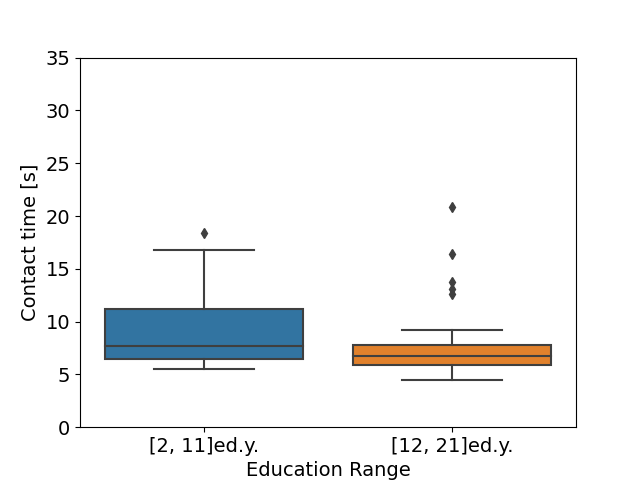}}
\subfigure[All, second age range]{\centering\includegraphics[scale=0.37]{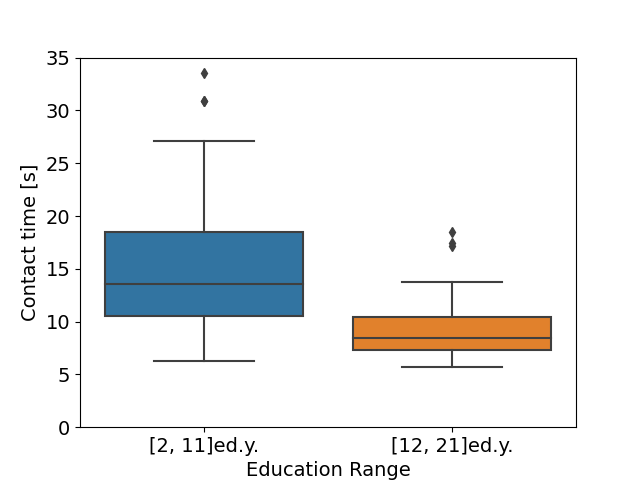}} 
\\
\subfigure[HC, first age range]{\centering\includegraphics[scale=0.37]{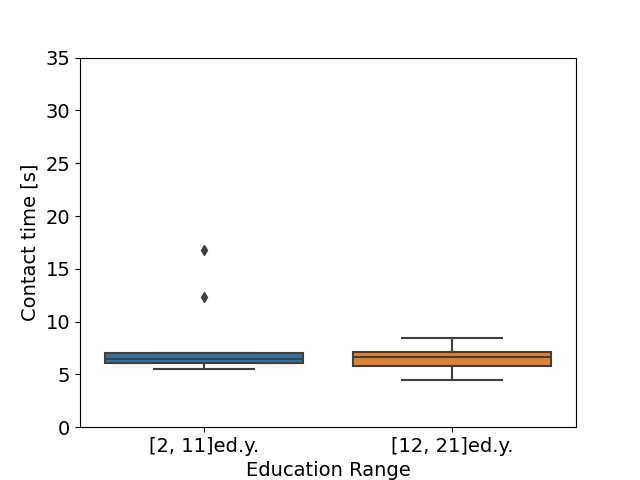}}
\subfigure[HC, second age range]{\centering\includegraphics[scale=0.37]{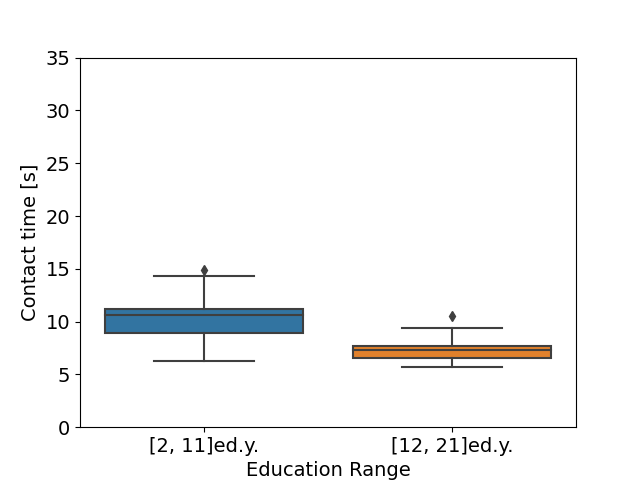}}
\\
\subfigure[PT, first age range]{\includegraphics[scale=0.37]{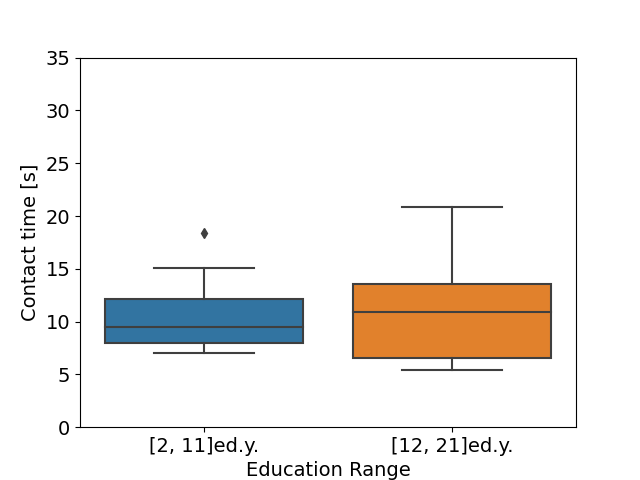}} 
\subfigure[PT, second age range]{\includegraphics[scale=0.37]{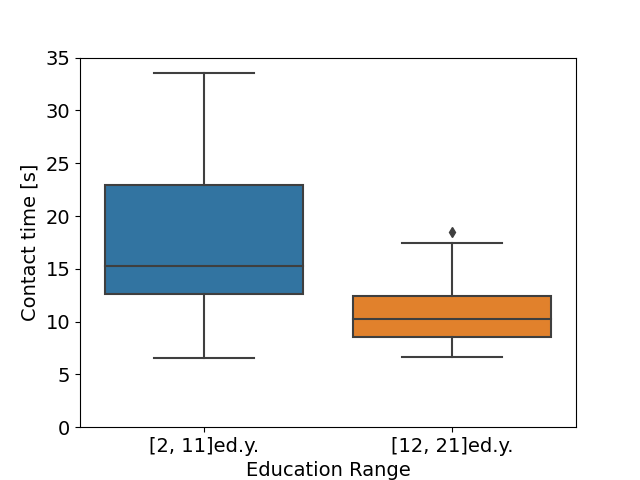}} 

\caption{Box plots showing how the contact time feature is related to age and education. The first age range is from 44 to 66 years old, while the second is from 67 to 88 years old.}
\label{fig:t23_c_time}
\end{figure*}

The above discussion suggests valuable features among the extracted ones. However, the results obtained from the first step procedure still require improvement for a classification system, especially in the medical field. To try and explain this behaviour, we analyzed our features deeply. First, we checked how the people in our dataset are distributed concerning personal features. Note that, from this point on, we refer to education level as the number of years of school attended by a person. This analysis was necessary as the dataset comprises people aged between 44 and 88 years old and with an education level which ranges from 2 to 21 years of school. We investigate whether these personal features could strictly influence handwriting tasks. Table \ref{tab:distributions} shows how people in our dataset are distributed according to age and education (school years). Both for age and education, we distinguished two ranges.
\\
Box plots in Fig. \ref{fig:t23_c_time} show how the contact time feature changes according to age, years of school and, obviously, the presence or not of Alzheimer's. These plots refer to the 23rd task, which outperformed all the others according to Table \ref{tab:ranking}. These representations are useful for comparing the distribution of a feature between the two education ranges considered for a particular group of people. In detail, the x-axis  shows the education range, the y-axis the contact time feature expressed in seconds, and every plot refers to a particular group of people (All, HC, PT) for a particular age range, where the first is $[44, 66]$ and the second is $[67, 88]$. 
\\
This figure highlights some evident trends:
\begin{itemize}
    \item Younger people are faster; if they have fewer years of school, they take more time and the deviation increases (Fig. \ref{fig:t23_c_time} (a) and (b));
    \item There's no variation between healthy people belonging to the first age range, independently from education (Fig. \ref{fig:t23_c_time} (c)); 
    \item Older healthy controls take more time than younger healthy controls, especially if they have a lower education level. It seems that education years are information that counts for elderly people (Fig. \ref{fig:t23_c_time} (c) and (d))
    \item The feature doesn't change so much for impaired people if they belong to the first age range, independently from the years of school (Fig. \ref{fig:t23_c_time} (e));
    \item Elderly patients show significantly different behaviour that depends on their education. Impaired elderly people take more time to complete handwriting tasks if they have fewer years of school (Fig. \ref{fig:t23_c_time} (f)).
\end{itemize}

Those findings hold for other features, too, mostly for the geometrical ones and those related to contact time and the number of lognormals. 
The relation between extracted features and personal information is extremely interesting.  This study allowed us to understand that a younger patient of AD could perform a task similarly or even better than an older healthy person with a fewer education level. Those findings can explain the performance obtained from the first step because there is an evident difference between young, healthy controls (c) and old patients (f). Furthermore, it decreases a lot between older healthy controls (d) and younger patients (e) or older ones but belonging to a major education range (f). There could be multiple reasons why healthy old people are not distinguished from younger patients. Older people, even if they don't suffer from AD, can have other impairments that affect their skills, and however, is known that a person's abilities deteriorate while getting older. Regarding younger patients, their execution depends on the stage of the illness, and they are probably more able to compensate for its effect concerning older people who lose control. It would be very interesting to know the stage of the disease of younger patients and to understand if they developed a compensation mechanism which makes them very similar to old healthy controls. 


\begin{figure}[htbp]
\caption{Distribution of people according to Education level and Age.}
  \begin{minipage}{0.45\linewidth} 
    \centering
   \begin{tabular}{c|c|c|c}
    \multicolumn{4}{c}{\textbf{Education level Distribution}} \\
    \hline
    \textbf{School years} & \textbf{Total} & \textbf{HC} & \textbf{PT} \\
    \hline
    $$[2, 11]$$ & 70    & 22    & 48 \\
    $$[12, 21]$$ & 104   & 63    & 41 \\
    \hline
    \end{tabular}%
  \end{minipage}
  \begin{minipage}{0.45\linewidth} 
  \centering
   \begin{tabular}{c|c|c|c}
    \multicolumn{4}{c}{\textbf{Age Distribution}} \\
    \hline
    \textbf{Age intervals} & \textbf{Total} & \textbf{HC} & \textbf{PT} \\
    \hline
    $$[44, 66]$$ & 69    & 50    & 19 \\
    $$[67, 88]$$ & 105   & 35    & 70 \\
    \hline
    \end{tabular}%
  \end{minipage}
  \label{tab:distributions}
\end{figure}

 \section{Conclusions and future work}
\label{sec:conc}

Alzheimer's disease is a progressive impairment which also affects handwriting, and it has no cure, so it is necessary to have an early diagnosis. In this research, we used a classification system based on ML, Ensemble and combining rules to distinguish between patients and healthy control through features extracted by applying the sigma-lognormal model to several handwriting tasks. The findings are interesting, though the results are still insufficient for a diagnostic aid system in the medical field. The best result from ML algorithms was achieved with an accuracy of 78.41\% on the 23rd task from RF. Regarding the stacking ensemble, the best performer was XGB, with an accuracy of 76.29\%. The majority vote combining rule obtains an 82.5\% of accuracy, combining the predictions from the first three tasks of a list where tasks are ranked according to their predictive ability. 
We discovered that the extracted lognormal features are useful in studying handwriting, particularly its dynamic and fluency. The results obtained are not enough for a medical problem, so we tried to find an explanation:
\begin{itemize}
    \item In our dataset we noticed that in some cases people belonging to the control group took a lot of time to execute some tasks, which is not expected. In other cases, people performing a task didn't follow strictly the requirements. Given this, we noticed that the velocity profile given by the sigma-lognormal model wasn't able to enhance the difference between HC and PT enough;
    \item Handwriting and the features are related to age and education. While it's easy to find differences between young HC and old PT, this difference strongly decreases when comparing old HC and young PT. 
\end{itemize}
This research has room to be improved, also, we mean to investigate more on our findings. We want to understand how to distinguish between older HC and younger PT, why they are so similar, and which technique or acquiring system we should consider. An interesting path that we mean to follow is to find a way to combine signal processing features with personal ones and understand if it is possible to measure or study how a person tries to cope with AD symptoms while writing and how he compensates. Future work will also focus on applying our system to different disease datasets and understanding if we obtain the same conclusions regarding results and discovered relations.

\subsubsection*{Acknowledgements.\\}
This work has been supported by the Spanish project PID2021-122687OA-I00 /AEI /10.13039 /501100011033 /FEDER, UE \\
\\
The research leading to these results has received funding from Project “Ecosistema dell’innovazione - Rome Technopole” financed by EU in NextGenerationEU plan through MUR Decree n. 1051 23.06.2022 - CUP H33C22000420001

\bibliography{IGS23}


\end{document}